**Domain adaptation of large language models for geotechnical applications**

Lei Fan[a,*], Fangxue Liu[a,b], Cheng Chen[c]

[a] Department of Civil Engineering, Design School, Xi'an Jiaotong-Liverpool University, Suzhou, China
[b] Department of Computer Science, University of Liverpool, Liverpool, L69 3BX, UK
[c] School of Civil Engineering and Architecture, China Three Gorges University, Yichang 443002, China
*Corresponding Author (Email: Lei.Fan@xjtlu.edu.cn)

**Abstract:** The rapid advancement of large language models (LLMs) is transforming opportunities in geotechnical engineering, where workflows rely on complex, text-rich data. While general-purpose LLMs demonstrate strong reasoning capabilities, their effectiveness in geotechnical applications is constrained by limited exposure to specialized terminology and domain logic. Thus, domain adaptation, tailoring general LLMs for geotechnical use, has become essential. This paper presents the first review of LLM adaptation and application in geotechnical contexts. It critically examines four key adaptation strategies, including prompt engineering, retrieval-augmented generation, domain-adaptive pretraining, and fine-tuning, and evaluates their comparative benefits, limitations, and implementation trends. This review synthesizes current applications spanning geological interpretation, subsurface characterization, design analysis, numerical modeling, risk assessment, and geotechnical education. Findings show that domain-adapted LLMs substantially improve reasoning accuracy, automation, and interpretability, yet remain limited by data scarcity, validation challenges, and explainability concerns. Future research directions are also suggested. This review establishes a critical foundation for developing geotechnically literate LLMs and guides researchers and practitioners in advancing the digital transformation of geotechnical engineering.

**Keywords**: large language model; geotechnical engineering; geology; domain adaptation; artificial intelligence

## 1. Introduction

Geotechnical engineering forms the foundation of modern infrastructure development and natural hazard mitigation. The discipline is inherently data-intensive and text-rich, relying heavily on the integration of diverse information sources, such as geological records, borehole logs, laboratory test results, design reports, and construction documentation (Wu et al., 2024). These materials require complex interpretation, expert judgment, and context-aware reasoning to support design calculations, modeling, and decision-making. Yet, due to the spatial heterogeneity and uncertainty of subsurface conditions, this process is often highly iterative, subjective, and labor-intensive. Consequently, the need to automate information extraction, interpretation, and synthesis has become increasingly urgent, particularly as project complexity and data volume continue to grow.

Over the past decade, the application of machine learning (ML) and deep learning (DL) in geotechnical engineering has advanced rapidly. Techniques such as support vector machines, random forests, and neural networks have been used to predict slope stability (Zhang et al., 2022), map landslide susceptibility (Chen and Fan, 2023a, 2023b), forecast landslide displacement (Weng et al., 2025), estimate bearing capacity (Karakaş et al., 2024), classify soils (Mitelman, 2024), calculate settlements (Kim et al., 2022), and model geotechnical risk (Liu et al., 2024). These approaches have undoubtedly improved prediction accuracy and computational efficiency. However, they remain primarily numerical and data-driven, struggling to handle the unstructured textual and semi-structured data that dominate geotechnical workflows. Moreover, their reliance on feature engineering and task-specific datasets limits scalability and interpretability—two factors that have become increasingly important for modern, data-integrated engineering practice.

In parallel, the rapid evolution of artificial intelligence (AI), particularly the advent of large language models (LLMs), has introduced a paradigm shift in how machines understand, process, and generate human language. Built on transformer architectures with self-attention mechanisms, LLMs such as OpenAI's GPT series and Google's Gemini can generate coherent, context-aware responses by leveraging knowledge from vast textual corpora (Chang et al., 2024; Raiaan et al., 2024). Enhanced through reinforcement learning with human feedback (RLHF) and increasingly capable of multimodal reasoning across text, image, and numerical data (Min et al., 2024), these models now represent a powerful new form of general-purpose cognitive automation.

For geotechnical engineering, a field characterized by document-heavy workflows, LLMs offer transformative potential. They can convert unstructured data into structured formats (Kumar, 2024), summarize complex design or geological reports (Ghorbanfekr et al., 2025), automate geotechnical interpretations (Xu et al., 2024), assist in design tasks (Xu et al., 2025a), support experiment planning (Yang et al., 2025), perform geoscientific analyses (Zhang et al., 2025a), and enable AI-assisted decision-making (Isah and Kim, 2025). In essence, LLMs provide an opportunity to bridge the long-standing gap between linguistic data and numerical modeling, transforming textual engineering knowledge into actionable computational intelligence.

Despite these promising developments, the use of LLMs in geotechnical engineering remains at an early and fragmented stage. Most existing studies are exploratory, focusing on narrow, proof-of-concept applications rather than systematic frameworks for integrating LLMs into engineering workflows. Critically, while the performance of LLMs in general domains is well established, their effectiveness in specialized technical contexts, where terminology, logic, and data structure deviate substantially from everyday language, depends heavily on how they are adapted to the target domain. Without such adaptation, general-purpose models risk misinterpretation of technical content, generation of factually incorrect outputs, or violations of engineering logic.

To address this challenge, domain adaptation, the process of refining general LLMs for specific disciplinary use, has emerged as a key research frontier. Yet, no comprehensive review has examined how these adaptation strategies are being implemented, evaluated, or optimized for geotechnical applications. The existing literature lacks a consolidated framework linking adaptation techniques with their geotechnical use cases and limitations.

This paper addresses that critical gap by:

- presenting the first review of domain adaptation of LLMs for geotechnical applications, covering representative studies published up to mid-2025.
- synthesizing how LLMs have been adapted and applied to core geotechnical tasks such as data extraction, interpretation, analysis, and design.
- offering a critical evaluation of current progress by assessing methodological advancements, identifying persistent challenges, and outlining opportunities for integrating LLMs within the broader geotechnical AI ecosystem.

The remainder of this article is organized as follows. Section 2 presents a bibliometric overview of current research activity. Section 3 outlines key techniques for domain adaptation of LLMs and compares their differences. Section 4 synthesizes adaptation strategies and applications within geotechnical workflows. Section 5 critically discusses open challenges and future directions. Section 6 concludes the article by summarizing insights and reflecting on the broader implications of LLMs for geotechnical engineering practice.

## 2. Bibliography screening and analysis

To identify relevant literature, this study employed two complementary sets of keywords: one focused on model-related terms (“large language model”, “LLM”, “artificial intelligence”, “AI”, or “GPT”) and the other on geotechnical-related terms (“geotechnical engineering”, “geotechnics”, “geoengineering”, “geology”, or “soil”). Using combinations of these complementary terms, a comprehensive search was conducted across multiple databases, including Web of Science, Scopus, arXiv, and Google Scholar. After a manual screening of abstracts, 31 studies were identified that explicitly address the application of LLMs to geotechnical tasks. The relatively modest number of publications can be attributed to the relative early stage of LLM technology, which has yet to achieve widespread adoption and evaluation within the geotechnical field. Nonetheless, the literature demonstrates a rapidly growing interest in this topic, with only 1 publication in 2023, 8 publications in 2024, and more than 20 in 2025 (up to the time of this survey). Given this accelerating trend, a review at the current stage is both timely and useful for assessing the state of progress and guiding future research directions in this emerging domain.

To further characterize the research landscape, we performed a bibliographic co-occurrence analysis of keywords extracted from the abstracts of the selected articles. The resulting network, shown in Figure 1, reveals a diverse and interdisciplinary research domain. The term “model” occupies a central position, acting as a conceptual hub that connects multiple thematic areas. Prominent terms such as “domain” and “GenAI” underscore the field’s growing emphasis on adapting LLMs for domain-specific geotechnical applications.

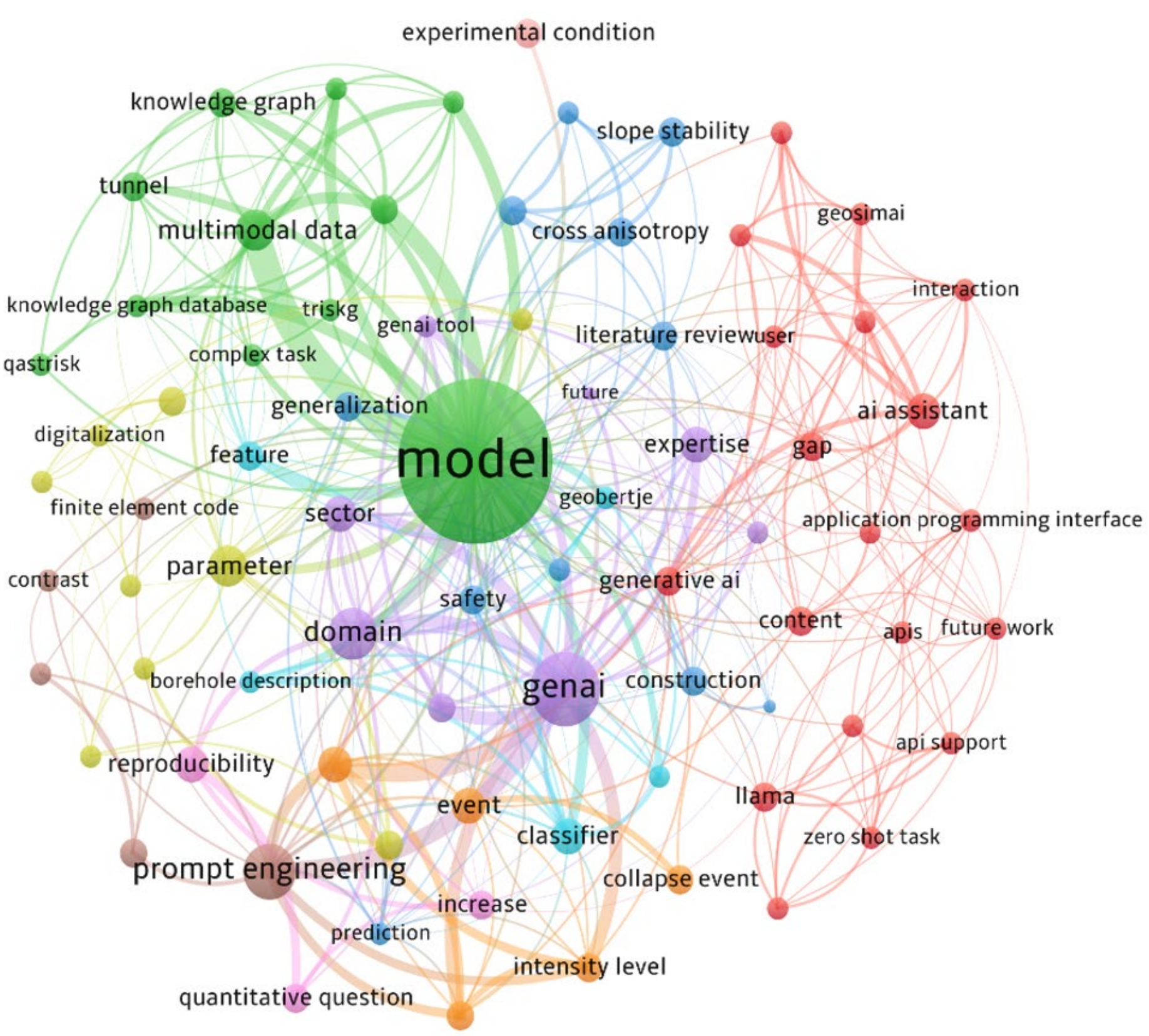

Figure 1. Bibliographic co-occurrence analysis based on the abstracts of surveyed articles concerning about the applications of adapted LLMs in geotechnical tasks.

Figure 1 also highlights several distinct thematic clusters. One major cluster centers on “multimodal data”, reflecting an emerging focus on integrating diverse data types, such as textual, visual, and numerical, to support more comprehensive geotechnical analysis. Another prominent cluster is anchored by "AI assistant" and "application programming interface", representing efforts to develop user-oriented tools that operationalize LLM capabilities for geotechnical interpretation and decision support. The co-occurrence of terms such as "gap" and "future work" within this cluster indicates ongoing challenges related to user interface design, tool usability, and effective domain adaptation.

A further notable cluster is built around "prompt engineering", which has become a key strategy for aligning LLM outputs with the specialized requirements of geotechnical tasks. Researchers increasingly employ structured prompts to guide LLMs in activities such as numerical reasoning, geological classification, and safety-critical assessments. This approach serves a dual purpose: it acts as both a lightweight domain adaptation technique and a human-in-the-loop mechanism that embeds expert knowledge without retraining the base model. Moreover, the focus on prompt design intersects with the broader concern of "reproducibility" in LLM research, underscoring the importance of transparent and standardized prompting practices in engineering applications.

## 3. Domain adaptation techniques of LLMs

The development of LLMs typically involves two main stages. In the pretraining phase, models learn language structure and general knowledge by processing large-scale text corpora (e.g., books, websites, encyclopedic data). However, these sources rarely focus on specialized geotechnical content. Analysis of common pretraining datasets reveals that geotechnical and geological content sparsely represented in typical training corpora (Lin et al., 2024), necessitating domain-specific adaptation for effective deployment. In the adaptation phase, models are tailored for specific applications, making them more suitable for practical use in fields like geotechnical engineering.

Four primary adaptation techniques have emerged as particularly relevant for geotechnical applications: prompt engineering, retrieval-augmented generation (RAG), domain-adaptive pretraining (DAPT), and fine-tuning, as illustrated in Figure 2. These are elaborated in Sections 3.1-3.4.

### 3.1 Prompt engineering: lightweight adaptation through intelligent querying

Prompt engineering remains the most accessible and least resource-intensive adaptation pathway. It operates exclusively at the input level, without altering internal model parameters, allowing practitioners to harness latent model capabilities through careful prompt design. Strategies range from explicit instruction-based prompting and few-shot demonstrations to structured reasoning scaffolds such as chain-of-thought and self-consistency prompting.

Empirical studies consistently demonstrate the significant sensitivity of LLM performance to prompt formulation. Brown et al. (2020) first revealed this phenomenon, and subsequent work in engineering contexts has substantiated it. For instance, Chen et al. (2024) demonstrated that task-specific instructional prompts could improve GPT-4's geotechnical problem-solving accuracy to 67%, compared to 28.9% and 34% with zero-shot and chain-of-thought prompting, respectively. Similarly, Li and Shi (2025) used example-based prompting to generate realistic geological cross-sections, illustrating that domain-contextual examples can "anchor" the model's reasoning to physically plausible outputs.

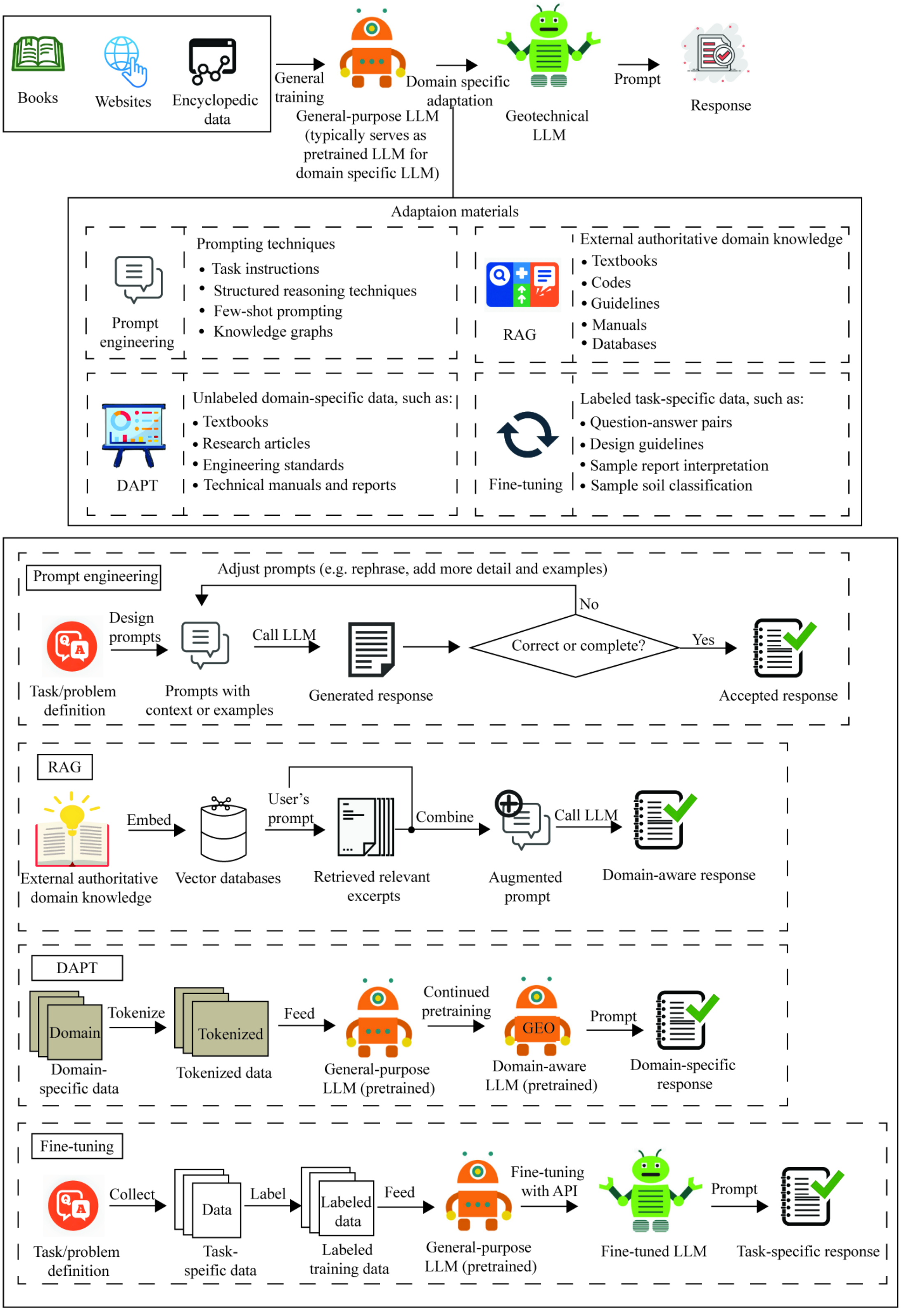

Figure 2. Typical domain adaptation techniques for LLMs

More sophisticated paradigms such as tree-of-thought reasoning (Wu et al., 2024) have further mitigated hallucination rates by encouraging LLMs to explore multiple reasoning pathways before finalizing an answer, a process reminiscent of expert deliberation in engineering design.

A particularly promising enhancement involves knowledge graph augmented prompting. By dynamically integrating structured ontologies of geotechnical entities (e.g., soil types, geological formations, load conditions), these methods provide factual grounding during inference. Xu et al. (2024) demonstrated that ontology-based graph integration notably improved LLM's predictive accuracy to above 90% in predicting adverse geological conditions during tunnel construction.

Nevertheless, prompt engineering remains a heuristic craft rather than a systematic science. Its effectiveness depends on user expertise, iterative experimentation, and contextual sensitivity, which collectively limit reproducibility.

### 3.2 RAG: bridging static models and dynamic knowledge

RAG represents a more structured attempt to compensate for an LLM's inherent knowledge cutoff. By integrating an external retrieval mechanism, RAG systems can "consult" authoritative domain references in real time, ranging from design standards to soil property databases, before generating an answer (Lewis et al., 2020; Cuconasu et al., 2024). This hybrid architecture grounds the generative process in verifiable sources, thereby enhancing factual reliability and transparency.

Tophel et al. (2025) demonstrated that a GPT-4 model augmented with a curated repository of geotechnical resources achieved 95% accuracy on domain quizzes, significantly outperforming its non-RAG counterpart. This magnitude of improvement is particularly important for engineering applications, where interpretive errors may have safety implications.

Recent advances extend RAG beyond single-document lookups. Qian and Shi (2025) introduced a multihop-RAG system that synthesizes design code requirements across multiple regional standards for automated borehole layout planning, effectively performing multi-source reasoning akin to human experts. This evolution from static retrieval to dynamic synthesis marks a critical step toward explainable AI in geotechnical practice.

While the RAG framework provides a powerful mechanism for dynamically grounding model outputs, its effectiveness is constrained by the characteristics of the underlying knowledge base, such as their quality, completeness, and continual updates. In highly localized contexts, such as national geotechnical codes or site-specific reports, retrieval errors or outdated information can propagate directly into model outputs. Thus, future RAG research should prioritize domain-specific corpus curation, multilingual retrieval mechanisms, and uncertainty quantification to ensure robustness.

### 3.3 DAPT: building domain fluency

Unlike prompt engineering or RAG, which act at inference time, DAPT operates during the model's learning phase. DAPT is an unsupervised method designed to enhance a LLM's understanding of a specific domain by continuing its pretraining on domain-specific corpora (Gururangan et al., 2020). For geotechnical engineering, such corpora may include textbooks, technical manuals, design reports, and research papers. The objective is to immerse the model in the domain's literature so that it internalizes the terminology, language patterns, and implicit conceptual knowledge characteristic of geotechnical discourse.

Through extended exposure to domain-specific syntax and semantics, the model develops a form of "native fluency", recognizing expressions such as consolidation settlement or Mohr–Coulomb parameters not merely as isolated tokens but as meaningful engineering constructs. Ghorbanfekr et al. (2025) demonstrated this effect by applying DAPT to a BERT-based model trained on a geotechnical corpus, resulting in improved comprehension of technical contexts prior to fine-tuning.

DAPT can thus establish a geotechnics-aware language model capable of producing and interpreting text with greater accuracy and contextual sensitivity to the field. Even without task-specific prompting, such a model exhibits enhanced understanding within its target domain. The adapted model can then be deployed directly or further fine-tuned for downstream tasks, as discussed in Section 3.4. For example, Ghorbanfekr et al. (2025) pretrained a BERT-based model on an unlabeled geotechnical corpus before fine-tuning it for specific applications.

Although computationally intensive, DAPT offers a promising balance between lightweight prompt-based adaptation and full-scale supervised fine-tuning. It effectively imbues the model with a "domain accent", enriching contextual awareness without enforcing task-specific behavior. However, from its application perspective, its scalability remains limited by the availability of high-quality, machine-readable geotechnical information and by the considerable computational resources required for large-scale pretraining.

### 3.4 Fine-tuning: achieving high task-level accuracy

Fine-tuning involves further training a pre-trained LLM to adapt it to a particular domain or application, often using a labeled, task-specific dataset (i.e., supervised training). Compared to DAPT, which enhances general domain understanding, fine-tuning aligns model behavior with task-specific objectives. This allows the model to learn new behaviors by exposing it to examples relevant to the target task, such as question-answer pairs, soil classification labels, interpretations of geotechnical reports, or design guidelines.

The process begins with the creation of a labeled dataset comprising task-specific input–output pairs. The size of this dataset can vary widely: while broad natural language processing applications may require tens of thousands of examples, specialized geotechnical tasks often achieve satisfactory results with smaller datasets due to the domain's focused vocabulary and structured knowledge base. These labeled data are then used to update the parameters of a base LLM, leveraging its general linguistic knowledge as a foundation, via a fine-tuning interface such as an application programming interface (API) or open-source library.

Fine-tuning can be conducted either on commercial cloud platforms (e.g., OpenAI) or locally using open-source frameworks (e.g., HuggingFace Transformers). Cloud-based fine-tuning offers ease of use and scalability but may raise data privacy concerns for proprietary geotechnical information. Conversely, local deployment demands greater computational capacity but ensures data confidentiality.

Once trained, the fine-tuned model demonstrates improved accuracy, consistency, and interpretive depth in processing domain-specific inputs. A representative example is GEOBERTje (Ghorbanfekr et al., 2025), which combines DAPT and fine-tuning for building this domain-adapted model. The fine tuning relied on approximately 2,500 labeled lithology classification samples. This two-stage approach achieved state-of-the-art performance, surpassing both traditional machine learning models and general-purpose LLMs, including ChatGPT-4. These results underscore the synergistic benefits of combining unsupervised domain adaptation with supervised task-specific training for specialized engineering applications.

Recent advances in parameter-efficient fine-tuning, such as adapter layers (Houlsby et al., 2019) and low-rank adaptation (LoRA; Hu et al., 2021; Dettmers et al., 2023), have made this process more accessible by updating only a small subset of model parameters. These methods reduce computational cost while retaining much of the performance improvement.

As geotechnical engineering increasingly depends on codified standards and empirical correlations, fine-tuned models hold potential as adaptive knowledge assistants. Realizing this potential, however, will require continued progress in data governance, model explainability, and validation practices.

### 3.5 Comparison of augmentation techniques

As presented in Table 1, the four adaptation methods form a continuum balancing cost, complexity, and performance. Prompt engineering offers immediacy and flexibility but often lacks depth and robustness. RAG improves factual grounding by integrating external sources, yet its reliability depends heavily on the quality and completeness of the retrieval corpus. DAPT cultivates domain fluency and contextual understanding but requires large-scale computational resources. Fine-tuning, while delivering the highest task precision, demands substantial computational resources and labeled datasets.

Table 1: Comparison of different domain-adaption methods

| Technique | Adaptation Level | Data Requirement | Computational Cost | Performance Gain | Key Advantages | Key Limitations |
|---|---|---|---|---|---|---|
| Prompt Engineering | Inference-level | None or minimal (few examples) | Very low | Moderate | Rapid deployment; low barrier to entry | Sensitive to prompt design |
| RAG | Inference-level | Moderate (curated text corpus) | Low - moderate | High | Enhances factual grounding and traceability | Requires corpus updates; retrieval errors |
| DAPT | Pretraining-level (unsupervised) | High (large unlabeled corpus) | High | High | Deep domain understanding; no labels needed | Limited domain corpora |
| Fine-Tuning | Training-level (supervised) | High (labeled dataset) | High | Very high | Precise domain alignment; strong downstream accuracy | Labeling cost; data privacy; potential overfitting |
| Hybrid | Multi-stage | Very high | Very high | State-of-the-art | Contextual grounding, domain fluency, task specialization | Complex integration; multidisciplinary expertise |

In practical applications, hybrid strategies are emerging as the most effective means of domain adaptation. For instance, combining DAPT with RAG can merge deep contextual fluency with reliable factual retrieval, while integrating fine-tuned adapters with optimized prompting can yield both accuracy and adaptability.

## 4. Adaptation and application of LLMs in geotechnical tasks

Table 2 presents an overview of the predominant application scenarios identified in the reviewed literature. It also summarizes representative geotechnical tasks and the corresponding domain-adaptation methodologies employed in these studies. Publications that do not specifically address geotechnical applications of LLMs are excluded. The table details key aspects such as the base models employed, adaptation methods applied, types of data used for adaptation, and the corresponding geotechnical problems addressed. Complementing Table 2, Figure 3 illustrates the distribution of adapted LLM applications across different geotechnical task categories, and the adoption frequency of different domain adaptation approaches.

Table 2 indicates that the ChatGPT series remains the most frequently used model across current applications. This predominance likely reflects both its status as the first widely available commercial LLM and its strong general performance. As shown in Figure 3, among domain adaptation techniques, prompt engineering emerges as the most commonly adopted strategy, significantly more prevalent than fine-tuning or RAG. This preference is largely due to its simplicity, accessibility, and minimal technical requirements, as it does not require domain specific corpora, labeled datasets, or specialized training infrastructure. For many practitioners, prompt engineering remains the only feasible method of model customization.

By contrast, fine tuning requires substantial computational resources, specialized expertise, and access to high quality domain data (often labeled) resources that are limited within the geotechnical discipline. These constraints explain its infrequent use in the reviewed applications. RAG approaches, while effective in integrating external knowledge sources, demand well-structured and clean corpora such as technical standards or design reports, as well as robust vector databases and retrieval pipelines. This added complexity has confined RAG primarily to experimental or research environments rather than routine engineering practice.

Figure 3 shows that research attention is fairly balanced across most task areas. Comparatively, adapted LLMs are particularly popular for hazard and risk assessment tasks, while site-planning tasks have received comparatively limited focus. Detailed discussions of these geotechnical applications are provided in Sections 4.1–4.6. These reviewed applications also reveal two distinct pathways for adapting LLMs to geotechnical contexts: (i) domain specialization (e.g., Wickramasinghe et al., 2024; Ghorbanfekr et al., 2025), and (ii) hybridization, which integrates LLMs with structured ontologies or multimodal embeddings (e.g., Li and Shi, 2025; Wang et al. 2025; Xu et al., 2025b; Areerob et al., 2025). The first approach enhances linguistic precision, while the second broadens reasoning capacity. The convergence of these two pathways is likely to define the next stage of AI supported geotechnical tasks.

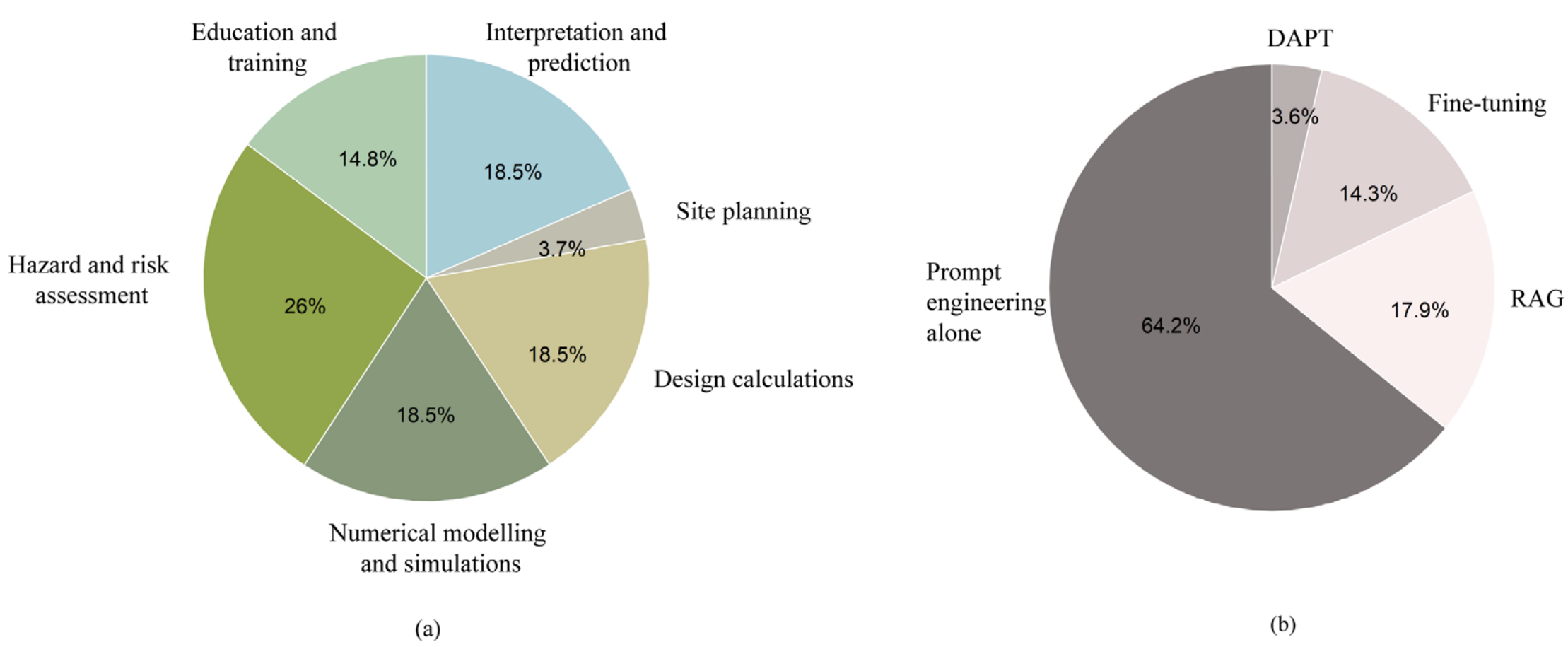

Figure 3. Statistics on geotechnical tasks and the associated domain adaptation in the surveyed literature: (a) Proportion of different geotechnical tasks using adapted LLMs; (b) Proportion of different domain adaptation techniques.

Table 2. Summary of representative approaches for domain adaptation of LLMs in geotechnical tasks

| Application scenarios | References (Year) | Adopted base models | Domain adaptation methods | Adaptation data | Applied tasks |
|---|---|---|---|---|---|
| Interpretation and prediction | Ghorbanfekr et al. (2025) | BERTje | DAPT and Fine-tuning | Geological borehole descriptions in Dutch | Interpret geological borehole descriptions into lithology classes |
| | Xu et al. (2024) | ChatGPT2 | Prompt engineering | Domain context, task instruction, and input statistics | Predict adverse geological conditions during tunnel construction |
| | Li and Shi (2025) | ChatGPT4 | Prompt engineering | Borehole data | Generate 2D visualization of geological cross-sections |
| | Zhang et al. (2025a) | ChatGPT-4o | Prompt engineering | Open data service of Mindat, pre-defined examples | Geoscience data interpretation and analysis |
| | Yang et al. (2025) | ChatGPT | Prompt engineering | Biochar characteristics identified by machine learning models. | Experiment planning in biochar immobilization of soil cadmium |
| Site planning | Qian and Shi (2025) | ChatGPT-4o | RAG | Site-specific design codes | Generate borehole sampling layouts and subsequent site characterization |
| Design calculations | Kim et al. (2025a) | ChatGPT4o, ChatGPTo1 | Prompt engineering | Site- and pile-specific data, API RP 2A standards | Generate codes to calculate the bearing capacity of piles |
| | Chai et al. (2025) | Not specified | RAG | Research papers, technical guidelines, and domain-specific software documentation | Settlement analysis and ground improvement |
| | Xu et al. (2025a) | Gemini-pro, Qwen-plus, ChatGPT-4, GLM-4 | Prompt engineering | Geotechnical instances, explicit instructions | Determine bearing capacity of piles and settlements |
| | Xu et al., (2025b) | ChatGPT-4o | RAG | Design examples with image and textual data, covering design assignments, solutions and reasoning steps | Design footings for meeting bearing capacity and settlement criteria |
| | Kumar (2024) | GPT-3.5-turbo | Prompt engineering | Data Interchange for Geotechnical and Geoenvironmental Specialists, and high-level function libraries | Soil classification, bearing capacity calculation |
| Numerical modelling and simulations | Bekele (2025) | Not specified | RAG | Theoretical and practical information on geomechanics, relevant datasets and target software for simulations | Generate scripts for geomechanical simulations, specifically for slope stability analyses with ADONIS, HYRCAN, FLAC and PLAXIS software. |
| | Kim et al. (2024) | ChatGPT | Prompt engineering | Conversational prompts include geometry, initial and boundary conditions. | Generate MATLAB codes for seepage flow analysis, slope stability, and X-ray computed tomography image processing of partially saturated sands. |
| | Kim et al. (2025b) | ChatGPT | Prompt engineering | Governing equations, boundary conditions, material properties, geometry, and numerical strategies etc. | Generate codes in FEniCS and MATLAB, respectively, to generate finite element models for hydro-mechanically coupled geotechnical problems |
| | Zhang et al. (2025b) | LLaMA | Prompt engineering | Textual construction knowledge and time-series sensor data | Encode temporal variations of grouting parameters |
| | Parsa-Pajouh (2025) | ChatGPT | Prompt engineering | Field measurement data | Generate synthetic geotechnical datasets for enhancing model training |
| Hazard and risk assessment | Wu et al. (2024) | ChatGPT | Prompt engineering | Slope photos with expert labelled descriptions and collapse risk assessment; HVSR curves and location coordinates at expert-selected sites | Slope stability assessment; Microzoning by seismic risk |
| | Wickramasinghe et al. (2024) | PHI3 and LLaMA3 | Fine-tuning | 5,000-entry Q&A dataset from national guidelines | Disseminate landslide-resilient construction knowledge to the public |
| | Hao et al. (2024) | ChatGLM-3b | Prompt engineering | News report, geographical locations | Detect ground collapse incidences from news. |
| | Isah and Kim (2025) | ChatGPT-3.5 | Prompt engineering | 42 published articles, covering diverse risk factors related to tunnel projects | Identify tunnel construction related risks |
| | Wang et al. (2025) | ChatGPT | Fine-tuning, and prompt engineering | Safety regulations and guidelines, historical safety incident reports, and construction site images | Identified construction hazards |
| | Kamran et al. (2025) | Gemini | Prompt engineering | Data Interchange for Geotechnical and Geoenvironmental Specialists (DIGGS), 93 rockburst cases | Predict geohazards (rockburst) during underground construction |
| | Areerob et al. (2025) | Alpaca 13B (LLaMA-2) | Fine-tuning and prompt engineering | Expert-based question answer pairs, covering types, causes, and observations of landslides; annotated images | Landslide disaster identification and risk assessment |
| Education and training | Tophel et al. (2024) | ChatGPT-4 | RAG | A dataset of textbook questions | Perform intelligent tutoring for solving technical problems in geotechnics |
| | Chen et al. (2024) | ChatGPT-4 | Prompt engineering | A dataset of textbook questions | Answer textbook-style geotechnical problems |
| | Kitaoka et al. (2024) | ChatGPT-4 | Prompt engineering | Key geological and topographic terms from survey reports | Generate tunnel construction related considerations |
| | Mahjour et al. (2024) | ChatGPT 3.5 | Prompt engineering | Contextual descriptions, numerical examples, and multiple-choice options | Evaluate risk and uncertainty in the Greenfield and Brownfield domains. |

### 4.1 Geological interpretation and prediction

LLMs are transforming the interpretation of geological data by linking unstructured descriptions with structured analytical representations. Traditionally, geotechnical engineers have depended on manual interpretation of borehole logs, core samples, and field reports, processes that are time-consuming and susceptible to inconsistency. LLMs now enable automated classification of soil types, identification of lithological transitions, and structured extraction of subsurface information from narrative text. These capabilities accelerate data processing and improve consistency, although expert validation remains essential to prevent misclassification and mitigate the risk of model hallucination.

The development of domain specific language models has further enhanced interpretive accuracy. For example, as shown in Figure 4, GEOBERTje (Ghorbanfekr et al., 2025) demonstrates how linguistic adaptation to a specific technical and cultural context, such as Dutch borehole logs, achieves superior results compared with general purpose models like GPT-4 and Gemma 2. GEOBERTje's performance advantage over Word2Vec based and rule-based methods highlights the strength of contextual embeddings relative to shallow linguistic representations in geological text analysis. This finding suggests that adaptation through fine tuning on domain corpora can be more effective than simply increasing model size, particularly for specialized geotechnical applications.

Beyond text-based interpretation, LLMs are also facilitating generative geological visualization. As shown in Figure 5, Li and Shi (2025) demonstrated that few shot prompting enables LLMs to reconstruct two dimensional geological cross sections from limited borehole data. Incorporating self-consistency mechanisms and knowledge guided examples helps to reduce reasoning drift, which remains a major limitation of general-purpose models. The generated cross sections were statistically consistent with expert interpretations, indicating the emergence of collaborative human–AI interpretive frameworks that may redefine early-stage site characterization.

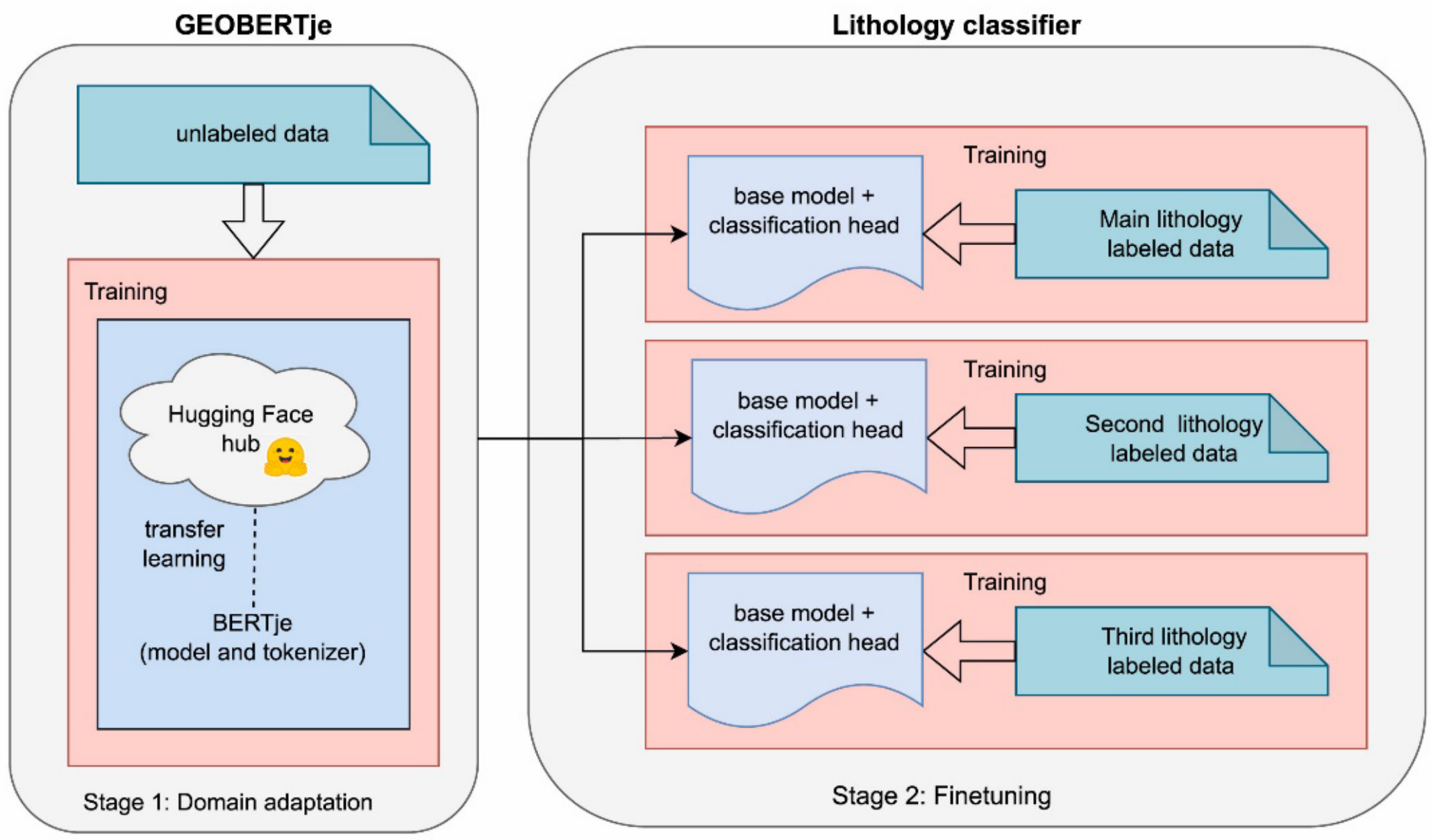

Figure 4. Task-specific adaptation workflow of GEOBERTje for the lithology classification (From Ghorbanfekr et al. (2025)).

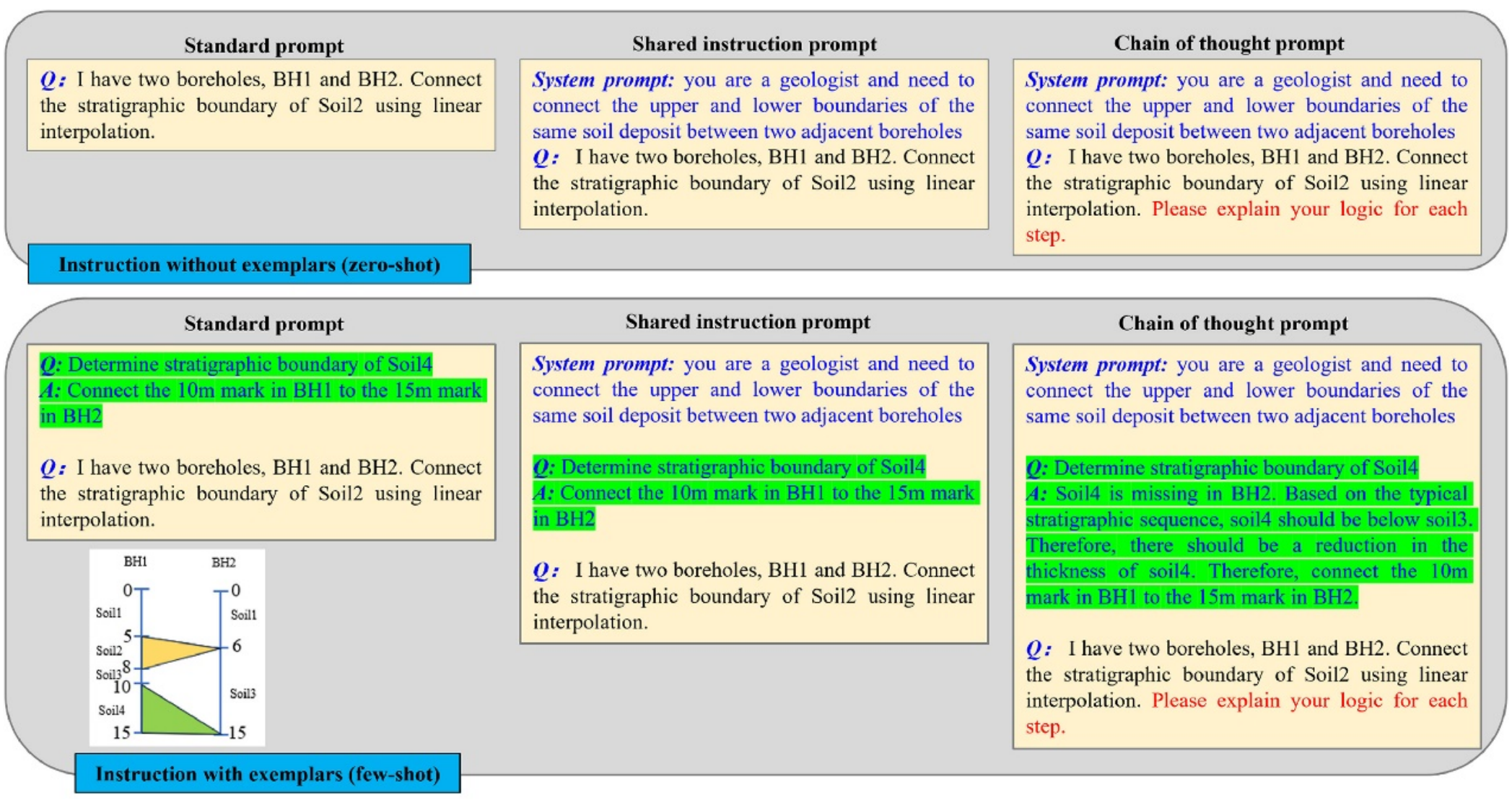

Figure 5. Prompting engineering for learning stratigraphic connectivity between boreholes (From Li and Shi (2025)).

In predictive applications, LLMs are increasingly integrated with structured knowledge systems. Xu et al. (2024) introduced GeoPredict-LLM, which combines multimodal data within a knowledge graph based embedding space. Its “prompt as prefix” strategy allows general purpose models to perform domain specific reasoning by merging symbolic knowledge representation with subsymbolic inference. This integration represents a shift from purely text driven analytics toward knowledge informed geotechnical intelligence.

### 4.2 Site planning

Site planning is an inherently complex process that integrates geotechnical, regulatory, and environmental considerations. LLMs are emerging as intelligent assistants capable of synthesizing these diverse dimensions to support semi-autonomous decision making.

Qian and Shi (2025) introduced the “Geologist” agent, depicted in Figure 6, which exemplifies this new direction. The system employs a multihop-RAG framework that dynamically retrieves and reasons across design codes, seismic zoning data, and soil classifications to produce borehole sampling layouts that comply with regulatory requirements. By embedding semantic reasoning within rule interpretation, the system bridges the often-overlooked gap between regulatory text and engineering design logic. The authors reported that Multihop-RAG achieved over 85% accuracy and improved information retrieval efficiency by 20%. This approach represents a conceptual shift from simple document retrieval toward knowledge orchestration, positioning LLMs not merely as information extractors but as interpreters of regulatory intent.

While these advances enhance efficiency and strengthen compliance assurance, they also highlight ongoing challenges related to model accountability. The Geologist agent depends on iterative collaboration between human experts and the model to verify and validate outputs, reflecting the current limitation of LLMs in providing transparent reasoning chains. Future research should focus on developing explainability frameworks and traceable prompting logic to ensure that automated site planning systems produce outputs that are both technically sound and defensible within regulatory contexts.

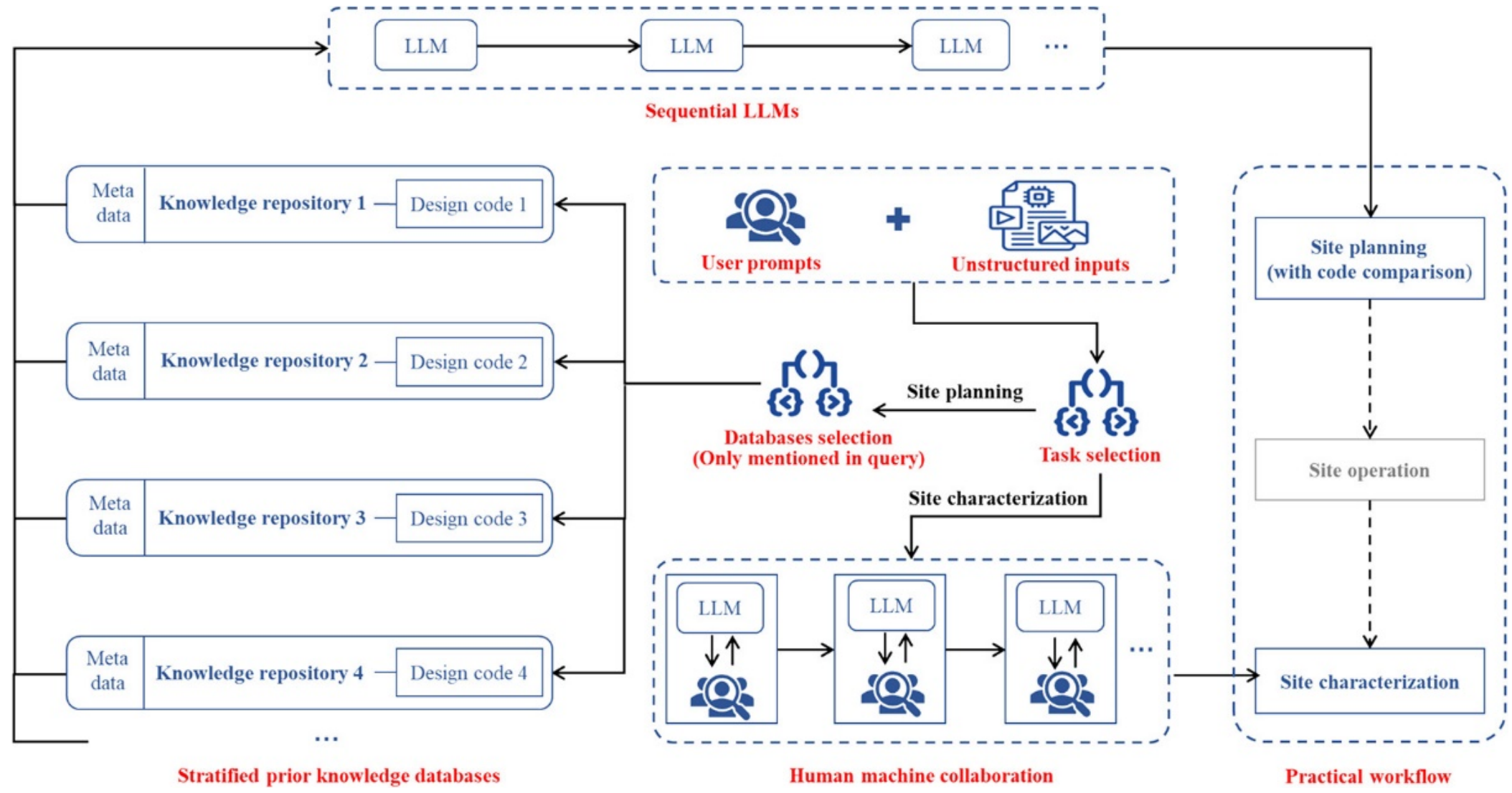

Figure 6. Framework of the "Geologist" agent paradigm for automated geotechnical site planning and characterization (From Qian and Shi (2025)).

### 4.3 Design calculations

Design calculations represent a critical frontier for the integration of LLMs, where linguistic reasoning meets quantitative precision. Recent studies have shown that, when supported by structured prompting or RAG systems, LLMs can assist in bearing capacity and settlement analyses with notable accuracy.

GeoLLM (Xu et al., 2025a) integrates natural language reasoning with mathematical toolsets to automate single pile design workflows, as illustrated in Figure 7. Its superior performance compared with smaller commercial models demonstrates the influence of model scale, while also revealing that model size alone does not guarantee improved accuracy without domain adaptation. Similarly, Kim et al. (2025a) found that ChatGPT performed reliably when generating code compliant computations based on API RP 2A standards, but produced errors when performing these calculations directly in natural language. This distinction highlights a fundamental boundary between linguistic fluency and numerical precision.

RAG-enhanced systems provide an effective alternative. Chai et al. (2025) showed that coupling LLMs with curated repositories of technical standards and design guidelines significantly improved interpretive accuracy. These results suggest that targeted context retrieval offers a more sustainable strategy for model adaptation than full retraining, particularly in regulatory environments that evolve rapidly.

Xu et al. (2025b) further advanced this approach through multi-GeoLLM, an agent based multimodal framework that integrates visual and textual data for foundation design. This system not only enhanced computational accuracy but also introduced a collaborative multi-agent structure that distributes tasks across specialized sub models, reflecting a broader trend toward modular architectures in artificial intelligence.

Collectively, these studies indicate that geotechnical design benefits most from hybrid intelligence. LLMs serve effectively as orchestration layers that translate natural language into structured computational instructions, while mathematical tools or deterministic solvers ensure numerical rigor. The next stage of innovation will likely involve closer LLM–solver co-design, where language models generate symbolic representations that are verified and refined through established computational methods.

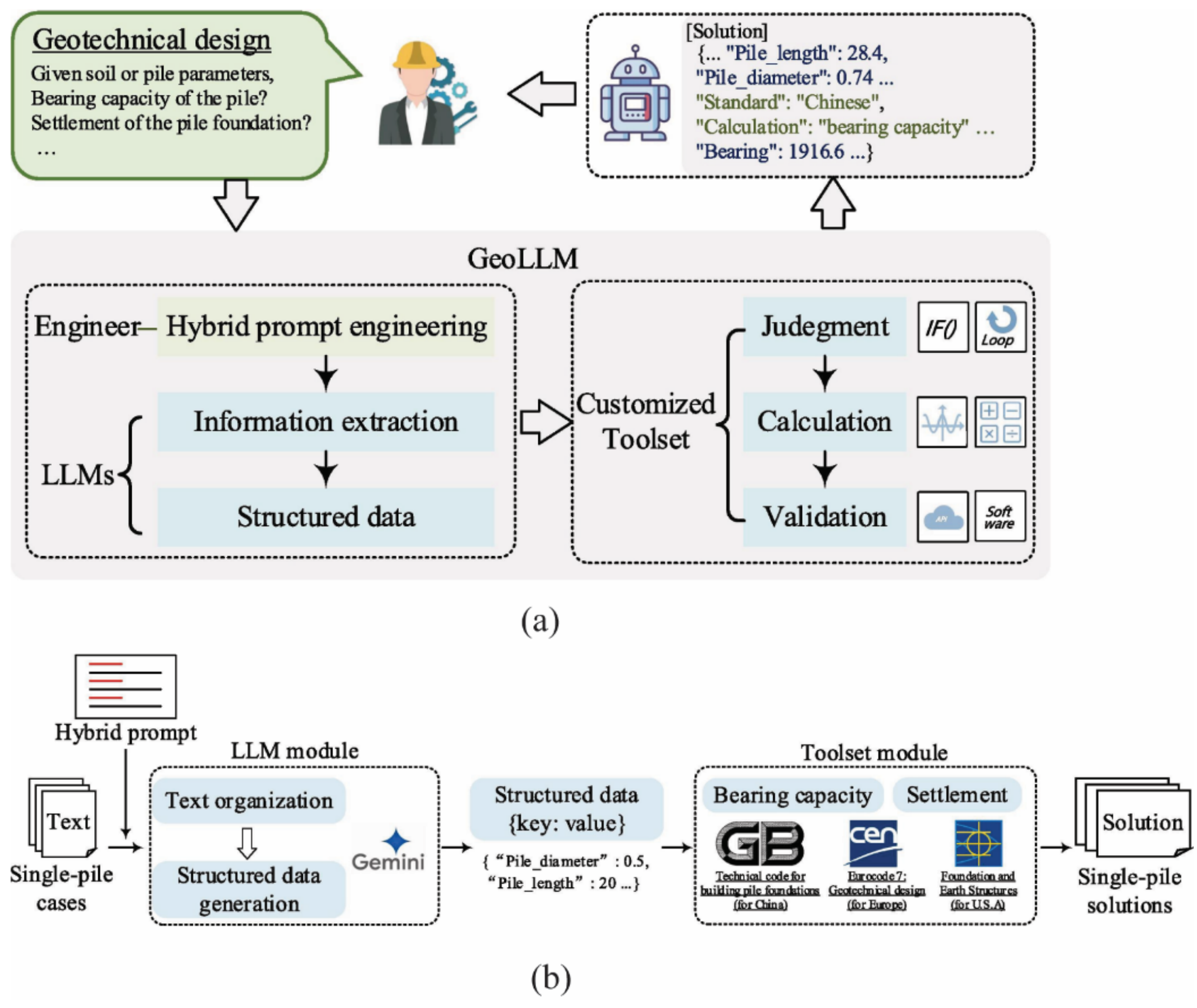

Figure 7. Architecture of GeoLLM: (a) framework for intelligent geotechnical design, (b) flowchart for single pile case (From Xu et al. (2025a)).

### 4.4 Numerical modelling and simulations

The integration of LLMs into numerical modeling offers significant potential to democratize simulation development by lowering the coding barrier and enhancing productivity. LLMs can translate natural language problem descriptions into executable scripts for finite element or finite difference analyses, a task traditionally reserved for experienced programmers. Figure 8 illustrates a representative workflow in which LLMs are used to generate executable code for finite element modeling.

Frameworks such as GeoSim AI (Bekele, 2025) illustrate how LLMs can serve as intelligent intermediaries between geotechnical engineering intent and numerical model syntax, significantly reducing the time required to set up analyses such as slope stability or seepage simulations. Building on this idea, Kim et al. (2024) showed that MATLAB scripts generated by ChatGPT for seepage flow and slope stability problems can yield results comparable to those produced by the commercial package GeoStudio SEEP/W. However, they also emphasized that achieving such accuracy still requires iterative human review and debugging. Extending this line of inquiry, Kim et al. (2025b) compared MATLAB codes and FEniCS-based finite element codes generated by ChatGPT and found the FEniCS implementations to be more effective, owing to the platform's higher level of abstraction and built-in computational capabilities.

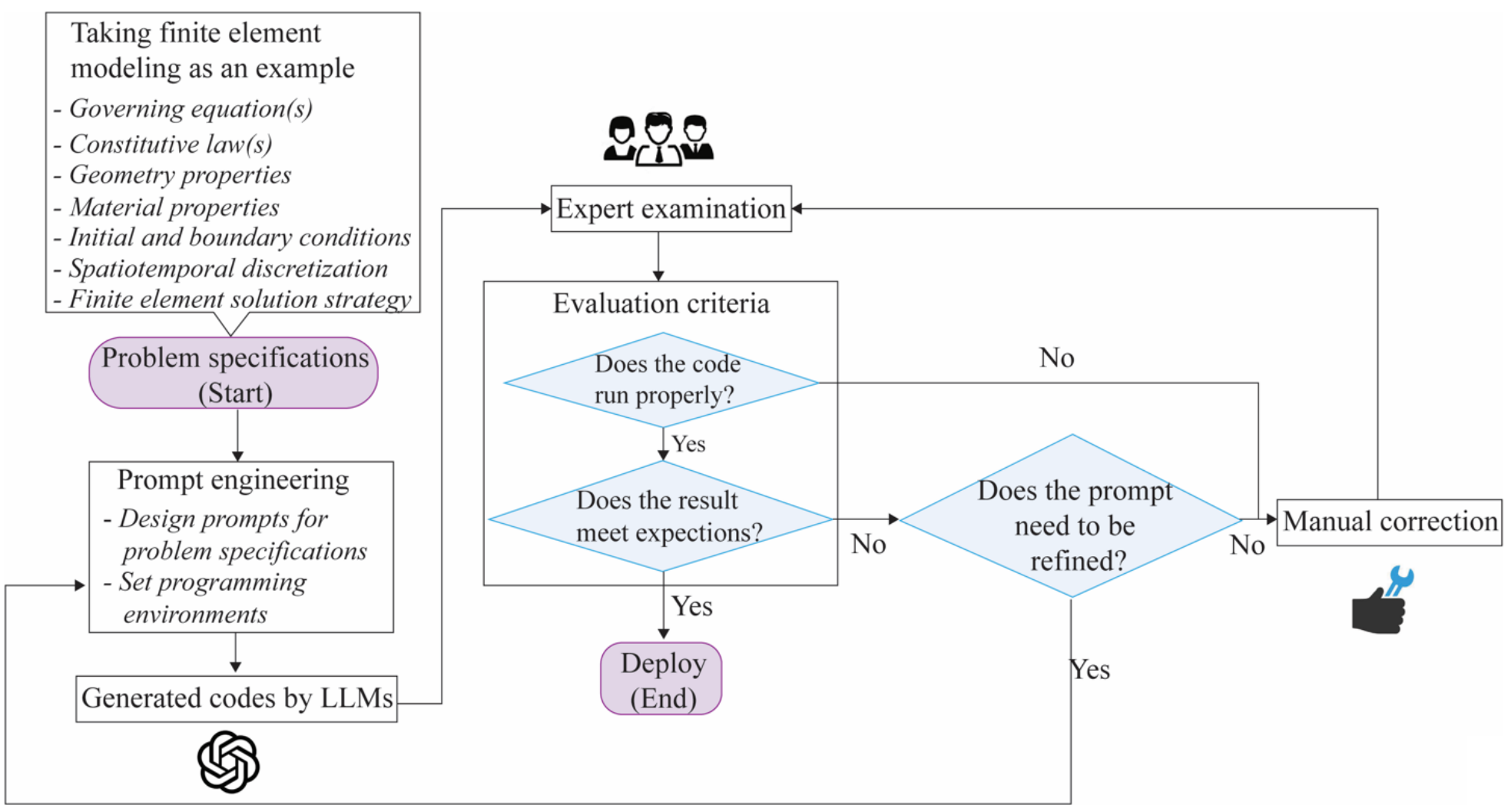

Figure 8. Representative workflow illustrating the use of LLMs to generate executable codes for geotechnical tasks, with finite element modeling as an example.

A comparison of these studies suggests that LLM effectiveness correlates with the level of abstraction in the modeling environment. Domain specific tools such as FEniCS require less detailed guidance, whereas general programming platforms such as MATLAB demand more structured prompting and expert oversight. Consequently, the most immediate value of LLMs lies in code scaffolding and workflow acceleration rather than in fully autonomous modeling.

Beyond code generation, hybrid systems are emerging that integrate LLMs with knowledge-based frameworks. KG-LLM (Zhang et al., 2025b) exemplifies this trend by embedding LLMs within physics informed knowledge graphs to support predictive simulations, such as estimating grouting parameters. These models demonstrate contextualized reasoning that combines linguistic pattern recognition with temporal sensor data, outperforming conventional regression-based approaches.

Another growing application involves synthetic data generation. Parsa-Pajouh (2025) demonstrated how LLMs can augment limited site-specific datasets to enhance model robustness. While promising, this approach introduces important challenges related to data provenance, bias propagation, and uncertainty management. Addressing these issues will be essential before synthetic data can be used in safety critical geotechnical applications.

### 4.5 Hazard and risk assessment

LLMs are increasingly being applied to geotechnical hazard assessment, particularly in tunneling and landslide prone environments. By integrating textual, visual, and numerical data, these models enable early hazard detection and more adaptive approaches to risk management. For example, Kamran et al. (2025) used Google's Gemini with tailored prompting to improve accuracy on the classification of rockburst intensity levels for underground construction.

Areerob et al. (2025) incorporated multimodal data into LLMs to enable expert-level analysis of landslide images, employing experts' question–answer pairs for landslide images for domain adaptation. Likewise, Wang et al. (2025) demonstrated that multimodal LLMs can relate visual construction scenes to safety protocols with high contextual relevance and reasoning fidelity. However, their accuracy remains sensitive to the specificity of prompts, a recognized limitation of generative models. Hao et al. (2024) adopted a complementary strategy by employing GPT models to extract information on urban ground collapse from public news sources. Although the recall rate (60%) was moderate, low precision (below 35%, due to frequent misclassification of other types of collapses) highlighted the need for robust post processing mechanisms to filter semantic noise in open-source data applications. Despite the novelty and utility of the strategy, the model's performance showed room for improvement.

More structured frameworks such as QASTRisk (Isah and Kim, 2025) have combined GPT 4.5 with tunnel risk knowledge graphs identify risks during the preconstruction phase by answering fact-based questions. As reported by them, this approach achieved a precision of 97%, a recall of 94%, and an F1-score of 95%, significantly outperforming traditional risk-identification methods. The integration of knowledge graphs supports semantic traceability and query based explainability, both of which are essential for regulatory acceptance and auditability.

The comparison between multimodal and knowledge-graph-oriented approaches reveals an important trade off. Visual models enhance situational awareness and contextual understanding, while symbolic systems offer transparency in reasoning and decision logic. The convergence of these paradigms could yield unified models capable of perceiving, reading, and reasoning, thereby redefining the role of artificial intelligence in geotechnical safety management.

In addition, the use of conversational agents for community engagement represents an often-overlooked dimension of LLM deployment. Wickramasinghe et al. (2024) demonstrated that chatbots can effectively communicate safety information to non-specialist audiences, extending the reach of geotechnical knowledge beyond the professional community and promoting broader public understanding of risk and resilience.

**4.6 Geotechnical education and training**

Education provides one of the most natural and immediate contexts for the adoption of LLMs in geotechnical engineering. Their conversational nature and adaptability make them valuable tools for personalized tutoring, content creation, and knowledge translation.

Survey results reported by Reddy and Janga (2025) indicate widespread use of LLMs in academic writing and coding tasks but relatively limited use in hands on analytical work. This reflects an ongoing gap between conceptual understanding and applied skill development. The strong performance of retrieval augmented GPT 4 in geotechnical quizzes (Tophel et al., 2024) demonstrates the potential of these models as intelligent tutoring systems. Chen et al. (2024) further showed that well designed instructional prompts can substantially improve accuracy, achieving 67% compared with 28.9% for zero shot prompting and 34% for chain of thought prompting. Nevertheless, persistent weaknesses in conceptual reasoning emphasize the continued need for human oversight.

The next stage of innovation lies in the development of adaptive learning ecosystems in which LLMs assess student understanding in real time and generate customized learning tasks rather than static answers. Embedding retrieval augmented LLMs into learning management systems could enable scalable and personalized geotechnical education.

In professional training, LLMs are also proving useful as knowledge mediators. Studies by Babu et al. (2025) and Kitaoka et al. (2024) have demonstrated their value in literature synthesis, report drafting, and interpretation of regulatory documents. Keyword based prompting techniques have been shown to improve response relevance while minimizing information leakage.

Despite these benefits, excessive dependence on LLMs in education may erode essential analytical rigor and creative problem solving, which remain central to geotechnical engineering practice. The challenge therefore extends beyond using LLMs effectively; it involves designing curricula that cultivate critical awareness of their limitations and promote responsible integration into engineering education (Fan et al., 2025).

## 5. Discussion

### 5.1 Domain adaptation challenges in engineering practice

For individual practitioners, prompt engineering provides a relatively accessible means of adapting LLMs to routine geotechnical tasks. However, developing effective domain-specific prompts requires both technical proficiency and deep subject-matter knowledge. As demonstrated in recent studies (e.g., Chen et al., 2024; Mahjour et al., 2024; Li and Shi, 2025; Reddy and Janga, 2025), this process can be time-consuming and error-prone, and often insufficient for tasks that demand complicated understanding of coupled physical processes or complex domain interdependencies (Tophel et al., 2025). Although RAG can improve prompt quality, verifying the accuracy and relevance of retrieved documents and integrating them effectively into the system requires substantial domain expertise, that many individual engineers may not possess. More advanced strategies, such as pretraining or fine-tuning, pose even greater barriers: curating high-quality geotechnical datasets and tailoring models to specialized tasks requires technical capabilities and resources that are typically beyond the reach of individual engineers.

Organizations may face similar or even amplified challenges, particularly for small and medium-sized enterprises. Implementing and maintaining a robust, context-aware retrieval infrastructure is resource-intensive (Zhang et al., 2025), especially when systems must support real-time, project-scale applications. Furthermore, pretrained and fine-tuned models may require continual updates as project requirements and regulatory standards evolve. This ongoing retraining process imposes significant operational costs (Wu et al., 2025). Additional concerns arise from dependence on cloud-based platforms (Reddy and Janga, 2025), including risks to data confidentiality, protection of intellectual property, and uncertainties stemming from the absence of explicit professional or regulatory guidelines governing AI-assisted engineering workflows. As a result, organizations face both technical and managerial challenges in deploying and sustaining LLM-based systems within geotechnical practice.

Therefore, there is a persistent gap in institutional readiness and human–AI integration. On one hand, systems such as the "Geologist" agent (Qian and Shi, 2025) and the multi-agent, multi-GeoLLM framework (Xu et al., 2025b) demonstrate that productive human–AI collaboration is technically achievable. On the other hand, individual practitioners in geotechnical organizations may not possess the dual expertise required to use such systems effectively: they must command geotechnical fundamentals while also understanding prompt engineering, contextual retrieval, and the inherent limitations of LLMs. As Babu et al. (2025) and Reddy and Janga (2025) emphasized, responsible deployment depends not only on technological capability but also on institutional approaches and user trainings. The field thus faces not merely technical challenges but institutional ones: without governance frameworks, validation protocols, and training mechanisms that equip engineers to critically assess AI-generated outputs, LLMs will remain confined to experimental domains rather than becoming fully integrated components of geotechnical engineering practice.

### 5.2 Technical challenges of LLMs in geotechnical tasks

Although recent developments demonstrate that domain adapted LLMs can perform a wide range of geotechnical tasks, their integration into practice remains constrained by several interrelated technical challenges. Evidence across the reviewed studies highlights three major areas of concern: factual reliability and numerical rigor, reproducibility, data scarcity and provenance.

The first and most critical challenge concerns factual reliability and numerical rigor. Numerous studies report that, despite producing fluent and contextually appropriate responses, LLMs continue to hallucinate and drift in reasoning, a pattern highlighted in the surveys of Reddy and Janga (2025) and Fan et al. (2025). In geological interpretation, for instance, Li and Shi (2025) observed frequent reasoning errors and stochastic behavior when their domain adapted models generated geological cross sections and stratigraphic transitions, even when the erroneous outputs appeared credible to non-experts. Similarly, Kim et al. (2025a) found that ChatGPT could provide more accurate, code-compliant answers when prompted within tightly structured constraints referencing API RP 2A standards, but it produced substantial numerical mistakes when asked to perform calculations in natural language. Xu et al. (2025a) further demonstrated that integrating language reasoning with mathematical toolsets within the GeoLLM framework significantly improved accuracy, reinforcing the point that linguistic fluency does not equate to engineering precision. Across these studies, domain adaptation improves performance relative to off-the-shelf models, but a notable performance gap remains in many cases when compared with conventional engineering methods.

The second common issue is reproducibility, which is central to engineering accountability. Mahjour et al. (2024), for example, documented inconsistent responses from ChatGPT when addressing complex questions involving risk and uncertainty, a problem echoed by Li and Shi (2025) in geological cross-section generation. To address such variability, researchers have explored fine-tuning domain-specific models such as "GEOBERTje" (Ghorbanfekr et al., 2025) or developing retrieval-augmented and knowledge-graph-based systems, such as "multi-GeoLLM" (Xu et al., 2025b) and "KG-LLM" (Zhang et al., 2025b), which link model reasoning to external, authoritative knowledge sources. Despite these advances, standardized mechanisms for documenting prompt context, retrieval provenance, and reasoning chains are still lacking in current studies. Without such mechanisms, LLM-generated outputs cannot meet the evidentiary standards required for design verification or engineering audit.

The third challenge arises from data scarcity, heterogeneity, and uncertain provenance. As emphasized by several studies (e.g., Mahjour et al., 2024; Wu et al., 2024; Li and Shi, 2025), geotechnical datasets are typically sparse, highly site-specific, and inconsistently formatted. For example, borehole logs, laboratory reports, field photos, and design charts are presented in mixed textual, tabular, and graphical forms, which are difficult to align within a single model input space. This fragmentation hampers both fine-tuning and robust retrieval-based adaptation. Given that geotechnical safety assessments depend on verifiable and traceable data, the provenance and validation of datasets must become an explicit component of any LLM-assisted workflow.

### 5.3 Future work

Building on current progress, future research should evolve beyond academic proof-of-concept demonstrations, toward systematic, verifiable, and ethically robust integration of LLMs into geotechnical engineering practice. The following research directions are proposed to guide this evolution.

Most existing studies have been conducted in academic settings, yet sustained progress will depend on coordinated efforts among research institutions, professional societies, and software developers. Such collaborations are needed to establish shared benchmarks, validation protocols, and data-governance frameworks, enabling LLMs to mature from isolated prototypes into reliable and auditable components of engineering workflows. Equally important is the development of standardized training programs by engineering firms and professional organizations to equip practitioners with the knowledge and skills necessary to implement AI responsibly in practice.

Future research should also prioritize the design of human-centered interfaces and governance structures that transform raw AI capabilities into trustworthy engineering tools. Work such as the Geologist agent (Qian and Shi, 2025) and the educational studies of Chen et al. (2024) and Reddy and Janga (2025) highlights the value of transparent, interactive systems that support iterative human–AI collaboration rather than passive acceptance of model outputs. Visual prompt builders, reasoning visualizations, and confidence indicators can enhance interpretability and user trust. Next-generation tools should enable dynamic interaction, allowing engineers to refine model outputs and examine explainable intermediate reasoning supplemented with quantified uncertainties.

Another priority is the co-design of LLM–solver architectures that guarantee numerical rigor. For instance, ITASCA's Particle Flow Code (PFC) has been widely applied in geotechnical research (Kodicherla et al., 2020), and its coding components can be supported by LLMs. Studies by Xu et al. (2025a) and Kim et al. (2025a) demonstrate that integrating LLMs with mathematical toolsets or domain-specific programming libraries reduces numerical errors relative to purely generative reasoning in LLMs. Future work should formalize these integrations through standard APIs that translate natural-language queries into executable formulations validated by established solvers.

Advances in traceable retrieval systems and knowledge-graph frameworks are also essential. Research on multi-GeoLLM (Xu et al., 2025b) and KG-LLM (Zhang et al., 2025b) shows that grounding model reasoning in structured knowledge improves both accuracy and reproducibility. Future studies should establish shared ontologies for soils, materials, laboratory tests, and design codes, and embed provenance metadata directly within model outputs to support auditability and regulatory acceptance.

Another major direction is the development of shared benchmarks, datasets, and evaluation standards. Current progress is hindered by fragmented data sources and a lack of uniform evaluation practices. Creating multilingual, multimodal benchmark datasets would enable objective performance comparisons across models. These datasets should be accompanied by open repositories of prompts, retrieval templates, and evaluation scripts to promote transparency and reproducibility.

Research should further advance multimodal data fusion, which has shown promise in geological visualization, safety assessment, and geohazard prediction (Wu et al., 2024; Li and Shi, 2025; Wang et al., 2025; Xu et al., 2025b; Areerob et al., 2025). Future work should develop adaptive alignment techniques that reconcile textual, visual, and numerical information while preserving geotechnical invariants, thereby approximating the integrative reasoning practiced by human experts. Furthermore, deep learning techniques originally developed for natural language processing but later extended to visual domains (Zhu et al., 2024a) can also be applied to enhance image interpretation.

Expanding on early work such as Parsa Pajouh (2025), synthetic data generation for domain adaptation represents another important research frontier, particularly in data-scarce contexts. Such synthetic datasets should be grounded in physics-based simulators and accompanied by metadata describing generation procedures, validation checks, and uncertainty bounds. Data augmentation techniques may also be employed to enrich the visual dataset, as is commonly done in deep learning approaches for visual data processing (Zhu et al., 2024b; Zhu et al., 2025).

The development of lightweight, edge-deployable LLMs also warrants attention for real-time and safety-critical applications. As illustrated in the risk-assessment frameworks of Kamran et al. (2025) and Isah and Kim (2025), on-site intelligence can improve hazard detection and response times. Compact, efficient models optimized for local operation would enable early-warning systems in connectivity-constrained environments such as tunneling and slope monitoring.

To date, most studies have focused on replicating traditional geotechnical tasks such as classification, design, and safety assessment. Future work should extend toward cross-disciplinary intelligence, leveraging the broad knowledge bases of LLMs to integrate geotechnical engineering with environmental, biological, and materials sciences. Emerging fields such as biogeotechnical and eco-hydrological modeling offer particularly promising opportunities for such interdisciplinary integration.

## 6. Conclusion

LLMs are beginning to redefine the landscape of geotechnical engineering, offering unprecedented capabilities for automating data interpretation, facilitating numerical modeling, supporting design and risk assessments, and enhancing knowledge transfer through intelligent tutoring systems. The reviewed literature demonstrates that adapted LLMs, particularly those in the ChatGPT series, have rapidly emerged as accessible and versatile tools within geotechnical workflows. Among existing adaptation strategies, prompt engineering currently dominates due to its simplicity and low computational requirements, while more sophisticated techniques such as fine-tuning, RAG, and DAPT are gaining traction as researchers pursue deeper domain integration.

Across the surveyed studies, domain-adapted LLMs have demonstrated measurable benefits in productivity, automation, and interpretive accuracy. When used to generate and execute computational codes rather than directly performing numerical calculations, these LLMs achieve significantly higher reliability, illustrating their strength as reasoning assistants rather than as autonomous analytical engines. These findings underscore the importance of appropriate task alignment where LLMs should supplement, not replace, established analytical workflows, ensuring that their generative capabilities are harnessed under the guidance of domain experts.

This review has shown that the frontier of LLM use in geotechnics is moving from task-specific experimentation toward system-level integration: coupling natural-language reasoning with numerical solvers, knowledge graphs, and multimodal data streams. Such hybrid intelligence systems promise a new era of more explainable and collaborative engineering, where humans and machines jointly construct and validate interpretations of the subsurface.

Nevertheless, the deployment of adapted LLMs in geotechnical engineering remains at an early stage and exploratory. Current challenges include limited availability of high-quality domain data, insufficient validation protocols, and persistent concerns about their factual reliability, numerical rigor, and reproducibility. Providing appropriate training for practitioners is also essential for the successful and ethical application of LLMs in engineering practice. Without systematic benchmarking, transparent evaluation frameworks, and adequate training, the robustness of LLM-driven decisions in safety-critical contexts remains uncertain.

Future research should therefore focus on building geotechnically literate LLMs through hybrid adaptation pipelines that combine various augmentation techniques and/or multimodal embeddings. Establishing shared datasets, open benchmarks, and standards for model auditing will be essential for fostering trust, accountability, and adoption in professional practice. Interdisciplinary integration with related domains such as hydrogeology, geophysics, and biogeotechnics will further expand their analytical reach.

**Authors contribution** L.F.: Writing–original draft, Writing–review & editing, Conceptualization, Investigation, Methodology, Validation. F.X.L.: Visualization, Investigation, Data Curation. C.C.: review & editing.

**Funding** The authors did not receive support from any organization for the submitted work.

**Data availability** No data was used for the research described in this article.

## Declarations

**Competing interests** The authors declare no competing interests.